\documentclass[11pt]{article}

\usepackage[final]{acl}

\usepackage{times}
\usepackage{latexsym}

\usepackage[T1]{fontenc}
\usepackage[utf8]{inputenc}

\usepackage{microtype}

\usepackage{inconsolata}

\usepackage{graphicx}

\title{Bridging Legal Interpretation and Formal Logic: Faithfulness, Assumption, and the Future of AI Legal Reasoning}

\author{Olivia Peiyu Wang \\
  University of California, Santa Cruz \\
  \texttt{pwang95@ucsc.edu} \\\And
  Leilani H. Gilpin \\
  University of California, Santa Cruz \\
  \texttt{lgilpin@ucsc.edu} \\
  }

\begin{document}
\maketitle
\begin{abstract}
The growing adoption of large language models in legal practice brings both significant promise and serious risk. Legal professionals stand to benefit from AI that can reason over contracts, draft documents, and analyze sources at scale, yet the high-stakes nature of legal work demands a level of rigor that current AI systems do not provide. The central problem is not simply that LLMs hallucinate facts and references; it is that they systematically draw inferences that go beyond what the source text actually supports, presenting assumption-laden conclusions as if they were logically grounded. This proposal presents a neuro-symbolic approach to legal AI that combines the expressive power of large language models with the rigor of formal verification, aiming to make AI-assisted legal reasoning both capable and trustworthy, thus reducing the burden of manual verification without sacrificing the accountability that legal practice demands.
\end{abstract}
 
\section{Problem Statement: Two Kinds of Correctness}
 
Legal professionals and AI researchers share the goal of deriving trustworthy, verifiable reasoning over legal text, but approach it through fundamentally different frameworks. Legal interpretation draws on background knowledge and contextual inference, while formal logic demands that all inferences be explicitly grounded in the text.
 
These frameworks are not different levels of rigor, but different modes of reasoning. A legally sound conclusion may be formally invalid because that norm is nowhere stated in the contract. This gap is largely invisible in legal AI research because most systems either mimic legal interpretation through language model training or enforce formal validity through symbolic methods, without acknowledging that the two regularly diverge. We argue that making this gap explicit, measurable, and addressable is one of the most important open problems in legal AI.

\section{Research Contribution: Measuring the Gap}
 
\paragraph{Dataset and re-annotation.}
We investigate this gap using ContractNLI \cite{koreeda2021contractnli}, one of the few benchmarks grounded in authentic contract language. Its original labels, produced by legally trained annotators, largely reflect legal interpretation. We do not dispute this. We re-annotate the same examples under a strict formal definition - a hypothesis $H$ is entailed by premise $P$ if and only if $P \wedge \neg H$ is unsatisfiable, contradicted if and only if $P \wedge H$ is unsatisfiable, and neutral otherwise.
 
The result is striking - there is a substantial proportion of label shifts, primarily from \textsc{Entailment} to \textsc{Neutral}. These are not errors. There are cases where the original conclusion depends on background legal knowledge or contextual assumptions reasonable for a lawyer to invoke but absent from the text. This predominant \textsc{Entailment}$\to$\textsc{Neutral} transition reveals a systematic gap between pragmatic legal interpretation and strict formal entailment. We further construct minimal pairs - for each divergent case, a minimally modified hypothesis that becomes formally entailed by supplying the missing assumption explicitly. This approach transforms an opaque gap into a tractable, analyzable object.
 
\paragraph{Experimental findings.}
We evaluate three paradigms across five LLMs (GPT \cite{agarwal2025gpt}, Claude \cite{anthropic2026sonnet46}, LLaMA \cite{llama3modelcard}, DeepSeek \cite{liu2025deepseek}, Qwen \cite{qwen2}): pure LLM classification, LLM reasoning over formal logical representations, and a neuro-symbolic pipeline combining LLM formalization with an SMT solver \cite{de2008proofs}. Formal structure improves accuracy, but \textbf{accuracy does not imply faithful reasoning}. High-performing models succeed by mimicking legal interpretation, including its implicit assumptions, rather than by reasoning formally. The SMT pipeline is more conservative, returning neutral classification whenever explicit grounding is lacking, and surfacing the gap rather than papering over it.
 
We identify three recurring failure modes 
\begin{itemize}
    \item \textbf{Assumption Injection:} the reasoning silently bridges gaps with unstated inferences.
    \item \textbf{Scope Laundering: }the reasoning presents informal conclusions as formally grounded.
    \item \textbf{Implicit Constraint Blindness: }the reasoning overlooks constraints present in formal representations
\end{itemize}
The dominant error across all models is \textsc{Neutral} $\to$ \textsc{Entailment} misclassification, reflecting systematic assumption injection. The \textsc{Entailment} $\leftrightarrow$ \textsc{Contradiction} confusions are rare, which indicates that the challenge is insufficient grounding, not logical inconsistency.
 
\paragraph{Minimal axiom framework.}
Furthermore, the \textsc{Neutral} classification does not necessarily need to be a dead end. Rather than stopping at a neutral classification, the system computes the \textit{minimal set of additional axioms} sufficient to shift the classification to \textsc{Entailment} or \textsc{Contradiction}, and presents them to a legal reviewer with a targeted question - does standard contract law or the contractual context implicitly establish this assumption? A lawyer answering \textit{yes} validates the implicit norm that formal logic cannot capture alone; a lawyer answering \textit{no} confirms the case is genuinely underspecified. In both cases, legal expertise is applied precisely where formal methods reach their limit. The complexity of the minimal axiom set is also diagnostic - many or complex axioms signal genuine interpretive difficulty, while few and simple axioms suggest confident automated classification.
  
\section{Research Agenda: Bridging the Gap}
 
The legal interpretation-formal logic gap is not unique to contract entailment. It arises wherever AI systems cite sources to support legal claims, which is increasingly common in LLM-assisted drafting, regulatory analysis, and litigation support \cite{magesh2025hallucination,reuters2026sullivan}. Prior work addresses \textit{fabricated citations} such as references that do not exist, a problem now largely tractable through retrieval verification \cite{agrawal2024language, cohan2019scicite}. The more difficult and consequential problem is \textit{stance misrepresentation} where the source exists and is retrieved correctly, but the claim overstates, understates, or mischaracterizes what the source says \cite{us_v_farris_2026}. A model trained to reason like a lawyer will routinely infer more from a cited source than it strictly supports, because that is what legal interpretation does. Between 50 and 80 percent of LLM responses in legal and medical domains are not fully supported by their cited sources \cite{agrawal2024language}, yet no formal detection framework exists for this failure mode.
 
We propose two contributions to address this - 
\begin{itemize}
    \item \textbf{A Benchmark Dataset: } The benchmark dataset should consist of LLM-generated legal and academic text annotated for stance misrepresentation at the claim level, using the three-way framework and minimal pair methodology from our research contribution.
    \item \textbf{Solver-in-the-Loop Training: } Rather than using LLM judges or human preferences as feedback, we will use a formal verification tool as a reward signal, teaching the model to distinguish formally supportable inferences from assumption-laden ones. When the solver flags an insufficiently grounded claim, it also computes the minimal axioms required to ground it, feeding directly into targeted human review at exactly the points where legal interpretation and formal grounding diverge.
\end{itemize}
 
\paragraph{Vision.}
The long-term goal is AI legal reasoning that operates transparently across both frameworks. It should be capable of legal interpretation when appropriate, formal grounding when required, and able to surface the minimal assumptions connecting the two for targeted legal review \cite{hildebrandt2018law,francesconi2023patterns}. We argue that understanding where and why current legal AI systems break is not a limitation but the foundation of this agenda. Only by honestly characterizing failure modes can we identify where AI assistance can be responsibly applied and build systems that proactively surface interpretive uncertainty rather than asking lawyers to verify conclusions after the fact \cite{dixon2025guidelines}. 

\bibliography{custom}

@inproceedings{koreeda2021contractnli,
  title={ContractNLI: A dataset for document-level natural language inference for contracts},
  author={Koreeda, Yuta and Manning, Christopher D},
  booktitle={Findings of the Association for Computational Linguistics: EMNLP 2021},
  pages={1907--1919},
  year={2021}
}

@article{hildebrandt2018law,
  title={Law as computation in the era of artificial legal intelligence: Speaking law to the power of statistics},
  author={Hildebrandt, Mireille},
  journal={University of Toronto Law Journal},
  volume={68},
  number={supplement 1},
  pages={12--35},
  year={2018},
  publisher={University of Toronto Press}
}

@article{francesconi2023patterns,
  title={Patterns for legal compliance checking in a decidable framework of linked open data: E. Francesconi, G. Governatori},
  author={Francesconi, Enrico and Governatori, Guido},
  journal={Artificial Intelligence and Law},
  volume={31},
  number={3},
  pages={445--464},
  year={2023},
  publisher={Springer}
}

@inproceedings{agrawal2024language,
  title={Do language models know when they’re hallucinating references?},
  author={Agrawal, Ayush and Suzgun, Mirac and Mackey, Lester and Kalai, Adam},
  booktitle={Findings of the Association for Computational Linguistics: EACL 2024},
  pages={912--928},
  year={2024}
}

@article{dixon2025guidelines,
  title={Guidelines for Judicial Officers: Responsible Use of Artificial Intelligence},
  author={Dixon Jr, Judge Herbert B},
  journal={The Judges' Journal},
  volume={64},
  number={2},
  pages={36--38},
  year={2025},
  publisher={HeinOnline}
}

@inproceedings{de2008proofs,
  title={Proofs and Refutations, and Z3.},
  author={de Moura, Leonardo Mendon{\c{c}}a and Bj{\o}rner, Nikolaj S},
  booktitle={LPAR Workshops},
  volume={418},
  pages={123--132},
  year={2008},
  organization={Doha, Qatar}
}

@article{agarwal2025gpt,
  title={gpt-oss-120b \& gpt-oss-20b model card},
  author={Agarwal, Sandhini and Ahmad, Lama and Ai, Jason and Altman, Sam and Applebaum, Andy and Arbus, Edwin and Arora, Rahul K and Bai, Yu and Baker, Bowen and Bao, Haiming and others},
  journal={arXiv preprint arXiv:2508.10925},
  year={2025}
}

@misc{anthropic2026sonnet46,
  author       = {Anthropic},
  title        = {Introducing Claude Sonnet 4.6},
  year         = {2026},
  url          = {https://www.anthropic.com/news/claude-sonnet-4-6},
  note         = {Accessed: 2026}
}

@article{llama3modelcard,
    title={Llama 3 Model Card},
    author={AI@Meta},
    year={2024},
    url = {https://github.com/meta-llama/llama3/blob/main/MODEL_CARD.md}
}

@article{liu2025deepseek,
  title={Deepseek-v3. 2: Pushing the frontier of open large language models},
  author={Liu, Aixin and Mei, Aoxue and Lin, Bangcai and Xue, Bing and Wang, Bingxuan and Xu, Bingzheng and Wu, Bochao and Zhang, Bowei and Lin, Chaofan and Dong, Chen and others},
  journal={arXiv preprint arXiv:2512.02556},
  year={2025}
}

@article{qwen2,
  title={Qwen2 technical report},
  author={Yang, An and Yang, Baosong and Hui, Binyuan and Zheng, Bo and Yu, Bowen and Zhou, Chang and Li, Chengpeng and Li, Chengyuan and Liu, Dayiheng and Huang, Fei and others},
  journal={arXiv preprint arXiv:2407.10671},
  year={2024}
}

@misc{us_v_farris_2026,
  title        = {United States v. John Farris, No. 25-5623 (6th Cir. 2026)},
  author       = {{United States Court of Appeals for the Sixth Circuit}},
  year         = {2026},
  month        = apr,
  day          = {3},
  howpublished = {\url{https://law.justia.com/cases/federal/appellate-courts/ca6/25-5623/25-5623-2026-04-03.html}},
  note         = {Decided April 3, 2026}
}

@misc{reuters2026sullivan,
  author       = {Freifeld, Karen and Scarcella, Mike},
  title        = {Sullivan \& Cromwell law firm apologizes for {AI} hallucinations in court filing},
  howpublished = {\url{https://www.reuters.com/legal/litigation/sullivan-cromwell-law-firm-apologizes-ai-hallucinations-court-filing-2026-04-21/}},
  year         = {2026},
  month        = apr,
  day          = {21},
  note         = {Reuters. Accessed: 2026-04-29},
  organization = {Reuters}
}

@article{magesh2025hallucination,
  title={Hallucination-free? Assessing the reliability of leading AI legal research tools},
  author={Magesh, Varun and Surani, Faiz and Dahl, Matthew and Suzgun, Mirac and Manning, Christopher D and Ho, Daniel E},
  journal={Journal of empirical legal studies},
  volume={22},
  number={2},
  pages={216--242},
  year={2025},
  publisher={Wiley Online Library}
}

\end{document}